\documentclass{article}

\usepackage{PRIMEarxiv}

\usepackage[utf8]{inputenc} 
\usepackage[T1]{fontenc}    
\usepackage{hyperref}       
\usepackage{url}            
\usepackage{booktabs}       
\usepackage{amsfonts}       
\usepackage{amsmath}        
\usepackage{nicefrac}       
\usepackage{microtype}      
\usepackage{lipsum}
\usepackage{fancyhdr}       
\usepackage{graphicx}       
\usepackage{multirow}       
\usepackage{gensymb}        
\graphicspath{{media/}}     

\title{Feature-Augmented Deep Networks for Multiscale Building Segmentation in High-Resolution UAV and Satellite Imagery
}

\author{
  Chintan B. Maniyar\\
  Center for Geospatial Research, Department of Geography \\
  University of Georgia \\
  Athens, USA \\
  \texttt{chintanmaniyar@uga.edu} \\
   \And
  Minakshi Kumar \\
  Photogrammetry and Remote Sensing Department \\
  Indian Institute of Remote Sensing \\
  Dehradun, India\\
  \texttt{minakshi@iirs.gov.in} \\
  \And
  Gengchen Mai \\
  SEAI Lab, Department of Geography and the Environment \\
  University of Texas at Austin, Austin, USA\\
  \texttt{gengchen.mai@austin.utexas.edu} \\
}

\begin{document}
\maketitle

\begin{abstract}
Accurate building segmentation from high-resolution RGB imagery remains challenging due to spectral similarity with non-building features, shadows, and irregular building geometries. In this study, we present a comprehensive deep learning framework for multiscale building segmentation using RGB aerial and satellite imagery with spatial resolutions ranging from 0.4m to 2.7m. We curate a diverse, multi-sensor dataset and introduce feature-augmented inputs by deriving secondary representations—including Principal Component Analysis (PCA), Visible Difference Vegetation Index (VDVI), Morphological Building Index (MBI), and Sobel edge filters—from RGB channels. These features guide a Res-U-Net architecture in learning complex spatial patterns more effectively. We also propose training policies incorporating layer freezing, cyclical learning rates, and SuperConvergence to reduce training time and resource usage. Evaluated on a held-out WorldView-3 image, our model achieves an overall accuracy of 96.5\%, an F1-score of 0.86, and an Intersection over Union (IoU) of 0.80, outperforming existing RGB-based benchmarks. This study demonstrates the effectiveness of combining multi-resolution imagery, feature augmentation, and optimized training strategies for robust building segmentation in remote sensing applications.
\end{abstract}

\keywords{deep learning \and building segmentation \and remote sensing \and Res-U-Net \and feature augmentation \and multiscale}

\section{Introduction}
With the increase in the availability of high spatial resolution remotely sensed data, automated feature extraction has become an important and popular research topic in the field of remote sensing \cite{dervisoglu2020comparison, cheng2016survey}. Feature extraction is a fundamental task in remote sensing technology and serves as a baseline for applications not limited to land-cover classification, hazard detection, mapping, and managing agricultural, urban, and geological resources \cite{cheng2016survey}. Since the last two decades, there have been continuous efforts in extracting features such as buildings, roads, trees, vehicles, etc. by using various ever-evolving methods.

Building segmentation and extraction has been one of the most sought-after research topics given its applications in urban planning, urban monitoring, urban change detection, etc. \cite{Li2018, Vakalopoulou2015}. Significant attempts are continuously being made on building segmentation, from using Digital Elevation Models with optical imagery to using high dimensional data processing methods with the availability of open-source high-resolution data and the advent of deep learning \cite{ait2023enhancing, Zhu2017}. One of the most prominent challenges in building extraction has been using spectral information-based classification algorithms such as Support Vector Machines (SVMs), decision trees, and random forests, as it becomes difficult to distinguish the buildings from the backgrounds \cite{chen2020object}. However, the recent development of deep Convolutional Neural Networks (CNNs) has resulted in a lot of studies using CNNs for building extraction by employing semantic segmentation, image fusion, and image classification approaches \cite{Ji2019, Zhao2018, el2016convolutional, Vakalopoulou2015, lari2007automated}. Apart from the inherent image processing-based challenges, the quality and volume of annotated and training data also poses as a major challenge in semantic building segmentation \cite{Zhang2023annotation}. The following sections comprehensively discuss the different methods that have been developed to extract buildings from remote sensing imagery.

\subsection{Non-learning based Building Segmentation Algorithms}

Building segmentation is one of the most important and prominent tasks for urban monitoring, management, and mapping. There has been a continuous evolution in the dedicated efforts of researchers worldwide for extracting buildings from remotely sensed aerial and satellite imagery \cite{Li2022}. Non-learning-based methods use spatial-spectral contextual information to link to the inherent characteristics of a building, such as corners, edges, texture, and spectral discrimination from other features. Based on this, building segmentation algorithms can be divided into three wide categories: 1) ones that primarily use spatial information to extract buildings, 2) ones that primarily use spectral information to extract buildings, and 3) ones that follow knowledge-based techniques to extract buildings.

\subsubsection{Spatial Information-based Building Segmentation}

Spatial information such as edges and corners have proven to be highly crucial in building segmentation. A study combined distinctive corners while estimating building outlines to extract buildings \cite{Cote2013}, but was unable to extract irregularly shaped buildings. To achieve accurate building segmentation for irregular shapes, edge detection-centric algorithms were proposed. Canny edge detection was widely used in conjunction with Hough transform \cite{Turker2015} and with Haar features \cite{Cohen2016} to develop algorithms for high-precision building boundary extraction, thereby also accounting for irregular shapes. Another study used line segments as guided features to extract buildings \cite{Wang2015rectangular} of predominantly rectangular shape. A consecutive study suggested using multi-directional gradients and built-up indices to further improve rectangular building segmentation performance \cite{Ngo2015}.

\subsubsection{Spectral Information-based Building Segmentation}

A prominent building segmentation challenge that has been consistent right from early methods based on morphological analyses to more recent methods based on complex analyses is discriminating between background and target, as well as identifying building objects from among spatially similar non-building features \cite{Eskandarpour2018}. To overcome these challenges, spectral cues such as color and geometrical cues such as line and shape have been incorporated for building segmentation from very high spatial resolution imagery \cite{Li2018}. In the early 2010s, a generic index called Morphological Building Index (MBI) was introduced to extract buildings from high-resolution satellite imagery, based on spectral morphological information \cite{Huang2011}. MBI successfully extracted buildings even with irregular footprint shapes, but it failed in shadowy regions and faced instance separation challenges for closely located buildings. Consequent to MBI, another study proposed a similar building segmentation index called the Morphological Building/Shadow Index (MBSI) specifically aimed at improving MBI for shadowy regions \cite{Huang2012}. Apart from these methods, several band ratioing techniques were used to propose various indices to extract built-up areas from multispectral satellite imagery \cite{Osgouei2019, as2012enhanced, Xu2008}.

\subsubsection{Knowledge-based Building Segmentation}

Knowledge-based methods establish rules based on both spatial as well as spectral methods. For instance, a study used a rule that rectangular buildings cast rectangular shadows and hence, instead of finding buildings, this study focused on finding shadows of a specific shape to locate and extract the buildings \cite{Ngo2015}. Another study also used the association of adjacency between buildings and shadows \cite{Shi2016buildingextraction}. A novel approach to extract buildings by first identifying and isolating the most interfering classes such as shadows and vegetation and then segmenting the image into ‘superpixels’ was suggested with the motivation to preserve precision along with accuracy \cite{Ngo2017}. One study used a complex database of rules derived from both spectral and spatial object-based features such as color, edge, corner, shape, size, texture, and local binary patterns to extract buildings with irregular shapes \cite{chen2018object}. Another study used the shadow information and sun-sensor geometry of that satellite image to extract building rooftops \cite{Gao2018}.

\subsection{Learning-based Building Segmentation Algorithms}

Deep learning techniques have the unique capability of extracting low-level as well as high-level features simultaneously \cite{ait2023enhancing, Xi2023, abdollahi2021integrating, Erdem2020, chollet2018deep}  from raw images, which makes them uniquely suited to improve image segmentation techniques in image processing. This contrasts with classic machine learning techniques such as SVM and random forest, where manual feature extraction needs to be done as a pre-processing step before the training process can begin. Using deep neural networks has recorded a great improvement in feature extraction/representation learning from remotely sensed images \cite{Ji2019, chen2018extraction, el2016convolutional, Zhao2016, Zou2015}. CNNs specifically have shown a great increase in the effectiveness of image feature extraction in terms of crisp segmentation, object detection, labeling, and extraction due to their powerful representation learning ability \cite{o2015, Qin2018}. CNNs, with their pixel-wise convolution and multiple layers of abstraction, are capable of extracting buildings despite their spectral homogeneity with the image background.

\subsubsection{Deep Learning Algorithms on Raw Spectral Bands}

Taking advantage of deep learning techniques, a lot of research studies have been carried out in the past decade to improve the existing benchmarks for building segmentation from high-resolution satellite as well as aerial imagery. Building-A-Nets is an adversarial network for robust extraction of building rooftops \cite{Li2018}. Multiple Feature Reuse Network (MERN) is a weight-sharing CNN that detects building footprints from high spatial resolution satellite imagery \cite{Li2018mfrn}. In the deep learning paradigm, Fully Convolutional Networks (FCNs) are best suited for object segmentation and extraction \cite{Hosseinpoor2020}. An FCN employs solely locally connected layers such as convolution, pooling, and upsampling layers while avoiding the use of dense layers so that it can naturally operate on inputs of arbitrary size and produces an output of corresponding spatial dimensions \cite{Long2015, Sun2018}. FCNs are widely used for building segmentation with the help of transfer learning. Some of the popular FCNs that are widely used in remote sensing applications for feature extraction, semantic segmentation, and image classification are VGG16 \cite{Simonyan2015}, ResNet \cite{He2016}, DeepLab \cite{chen2018encoder}, DenseNet \cite{Sherrah2016, Yang2018}, SegNet \cite{badrinarayanan2017segnet}, deconvNet \cite{Noh2015} and UNet \cite{Ronneberger2015}. Bittner proposed a Fully Convolutional Network (FCN) to predict buildings by combining high-resolution remotely sensed imagery with normalized Digital Surface Models (DSMs) \cite{bittner2018building}. Consequently, another FCN called DE-Net, was proposed for building segmentation with more emphasis on information preservation during the upsampling, downsampling, and encoding stages of the network \cite{Liu2019}. Moreover, deepening an FCN and using more spectral bands has also proved to improve building segmentation performance from high-resolution satellite images \cite{Mohanty2020}. 

An increasing number of studies also focus on building segmentation from UAV (aerial) images. One such study ensembles two deep networks, SegNet and U-Net, and reports significant improvement in building segmentation from UAV imagery \cite{Abdollahi2020}. Other studies use feature enhancement techniques such as dilated spatial pyramid pooling \cite{me2020increasing}, attention pyramid networks \cite{Tian2022}, multi-stage multi-task learning \cite{Marcu2018}, and channel attention mechanisms \cite{Pan2019} along with the existing deep learning networks to improve building segmentation results from UAV data. Multiple studies that implement different variants of the U-Net architecture is a testament to the fact that U-Net is one of the most suitable architectures for extracting densely located buildings \cite{ait2023enhancing, Yang2023, Erdem2020, khosh2020multiscale, Huang2018}. Being an encoder-decoder architecture, U-Net is naturally suitable for building segmentation tasks and has been used with high-resolution satellite imagery \cite{Pan2020, Qiu2023}, with RGB images that fuse spatial image features \cite{abdollahi2021integrating}, and in an attention network-based supervised prediction \cite{Xu2021}. Another effective architecture for building segmentation has been Res-U-Net, which is a combination of ResNet, an image classification state of the art, and U-Net. Res-U-Net has been preferred especially in cases where transfer learning is used for building segmentation \cite{Dixit2021, Yoo2022}.

\subsubsection{Deep Learning Algorithms on Derived Feature Bands}

Apart from these, more recent studies try to improve the building segmentation performance on RGB images by further adding more feature bands that are derived from the RGB channels. This is because working on high-resolution or very high-resolution imagery always means restricted spectral information, owing to the spatial-spectral trade-off in remote sensing \cite{Kadhim2016}. Hence, typically very high spatial resolution imagery only has RGB or up to NIR bands. As a result, typically, a Res-U-Net faces significant challenges with irregular shapes and boundaries as well as due to interference by vegetation class. One study on very high-resolution imagery examines how training on RGB images and additionally on feature bands can be helpful in such cases \cite{Yang2018}. This study proposed to use various feature bands to improve building footprint extraction, e.g., using Sobel to enhance edges, utilizing DSM to extract buildings from undulations, and employing vegetation indices to discriminate between buildings and vegetation. The results showed that learning on images with RGB channels along with additional feature bands performed better than solely learning on RGB images. Similarly, other studies also attempted deep learning-based building segmentation models by using secondary feature bands on benchmark datasets of ISPRS Vahinigen, Massachusetts Building Dataset, and Postdam \cite{boonpook2021deep, Liu2018semantic}. The results also indicated that using feature bands along with the RGB channels can yield better building segmentation performance with finer boundary and edge preservation.

\subsection{Objectives and Summary}

This research is an extension to one of our previous works in which we have proposed building segmentation from a single sensor RGB UAV dataset using a pre-trained deep network \cite{maniyar2021deep}. Buildings tend to show the maximum conflicts between the foreground and background, as compared to other features such as waterbodies and roads when being extracted using spatial-spectral contextual information \cite{chen2020object, Gao2019, Kaplan2017, Vakalopoulou2015, Momm2011}. Moreover, a single CNN typically faces challenges to extract buildings from multiple scales (spatial resolutions) of remotely sensed images, which restricts the effective usage of a CNN to only one spatial resolution (Martins et al., 2020). In the current study, we aim to advance building segmentation models by not only enhancing existing deep CNN-based architectures for multiscale building segmentation but also investigating appropriate guiding features in a spectrally restricted setting apart from the RGB channels to improve building segmentation. The main innovation and novelty of this study is the implementation of multiscale deep learning architecture and using not features beyond the RGB channels to segment buildings. This study makes the following specific contributions:
\begin{itemize}
    \item Curate a multiscale and multi-sensor RGB UAV and satellite imagery dataset with spatial resolutions ranging from 0.4m to 2.7m, for building segmentation
    \item Identify and investigate secondary features from RGB imagery to improve upon building segmentation performance using deep learning networks
    \item Suggest a deep learning architecture capable of learning building segmentation at multiple scales (spatial resolutions)
    \item Suggest training policies such as layer freezing-unfreezing, cyclical learning, and SuperConvergence to optimize training time and resource consumption during deep learning
\end{itemize}

\section{Materials and Methods}

\subsection{Study Area}

\begin{figure}
  \centering
  \includegraphics[width=01.0\linewidth]{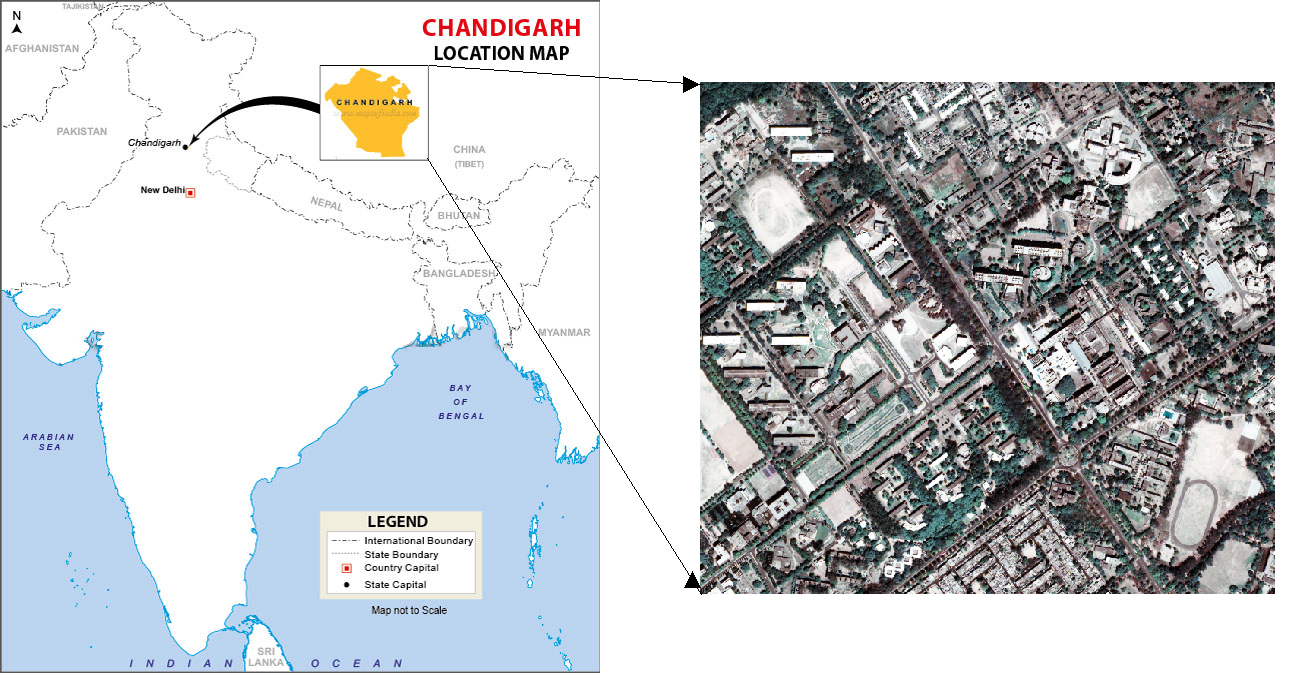}
  \caption{Inset map and snapshot highlighting Chandigarh City, India; the test site for this study.}
  \label{fig:inset-map-chandigarh}
\end{figure}

The training data of this study has been curated from multiple global datasets. Hence, the study sites in the training data span across cities in the United States, Austria, France, Eastern Asia, etc.
For the testing site, Chandigarh city of India (\autoref{fig:inset-map-chandigarh}), which is a union territory and serves as the capital city of two states namely Haryana and Punjab, was chosen. It is a well-developed city and flaunts an array of buildings in varying densities throughout its urban spread. It contains patches of recently developed very well-planned built-up areas and preserved patterns of primitive built-up areas that have not been changed. Moreover, it contains buildings of all shapes and sizes with intricate placement patterns and a wide spectral variety of rooftops. All these factors made it a desirable site to perform the testing for this study.

\subsection{Training Dataset Curation}

For the multiscale training purpose, this study curated a multiscale dataset from five open-source aerial and satellite datasets, namely:
\begin{enumerate}
    \item CrowdAI Mapping Challenge Dataset: It primarily focuses on the extraction of buildings under the general domain of object detection and object segmentation in computer vision \cite{Mohanty2020}. This dataset is derived from the SpaceNet dataset \cite{van2018spacenet} which contains orthorectified satellite imagery from the DigitalGlobe WorldView-2 and WorldView-3 satellites, containing images from up to five cities while covering at least 200 km2 area in each city;
    \item Open Cities AI Mapping Challenge Dataset: It is organized by the Global Facility for Disaster Reduction and Recovery (GFDRR), containing very high-resolution orthorectified aerial imagery at ~20cm spatial resolution in red, green, and blue bands, and over 790,000 building footprints from roughly 400km2 diverse African urban areas;
    \item Inria Aerial Image Labelling Dataset: It was curated to generalize semantic labeling methods for different urban environments \cite{Maggiori2017inriadataset}, comprises of high-resolution aerial RGB images from densely populated areas of Chicago and San Francisco in the US and the alpine towns of Austria;
    \item Massachusetts Housing Dataset: It was curated by Mnih et al. \cite{Mnih2013}, and comprised 151 aerial RGB images of Boston City at 1m spatial resolution;
    \item WHU Building Dataset: It was developed by the Group of Photogrammetry and Computer Vision at Wuhan University \cite{Ji2019}, and comprises aerial and satellite RGB images of the Eastern Asian regions at 2.7m, captured from QuickBird, IKONOS, and ZY-3.
\end{enumerate}

\begin{figure}
  \centering
  \includegraphics[width=01.0\linewidth]{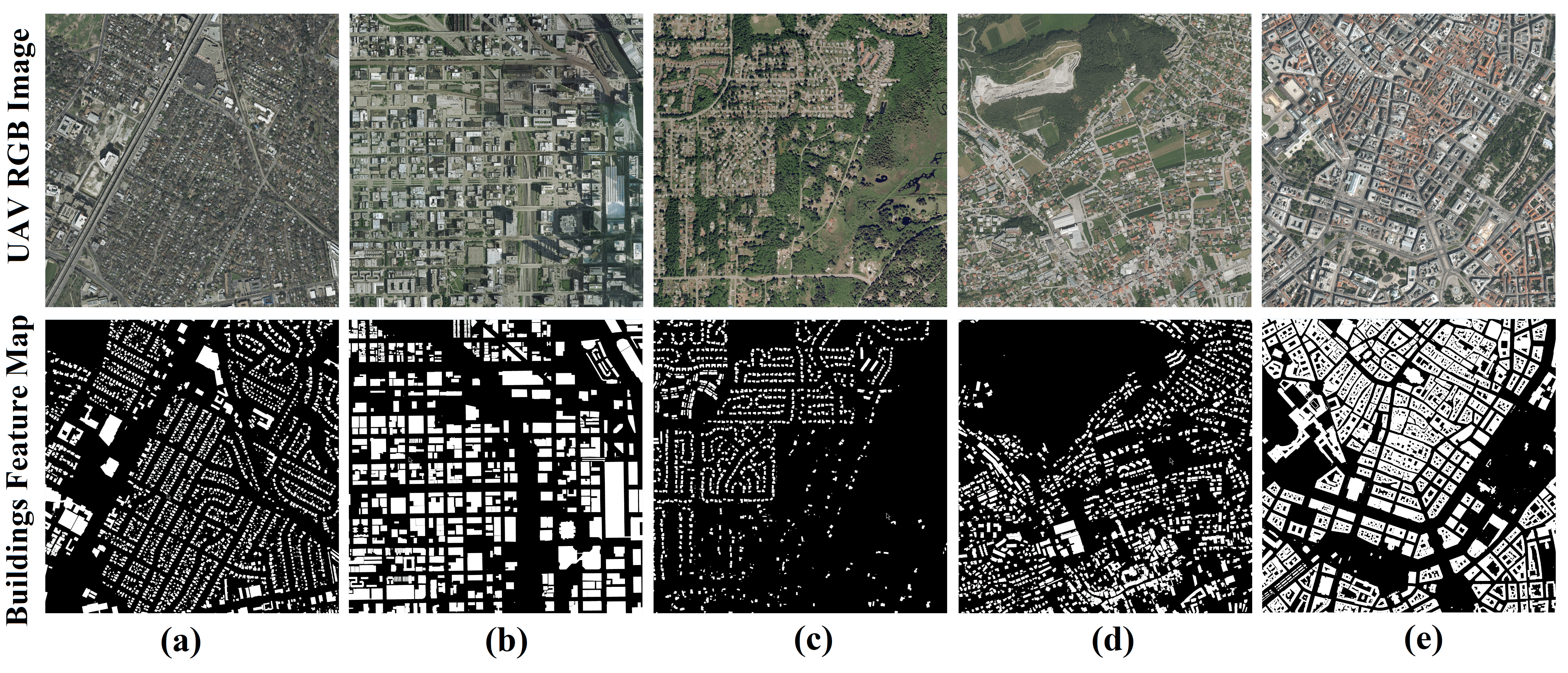}
  \caption{Image and label pairs from each of the five datasets: (a) Inria Image Labelling Dataset, (b) Wuhan Housing Dataset, (c) Massachusetts Building Dataset, (d) CrowdAI Mapping Challenge Dataset, and (e) Open Cities AI Mapping Challenge Dataset}
  \label{fig:trainig-samples}
\end{figure}

This study only uses RGB channels with the motivation that if good results can be achieved on RGB, extending the dataset beyond the RGB channels could exhibit even better results. \autoref{fig:trainig-samples} shows the nature of the dataset in terms of the images and the corresponding building labels by using sample images from the Inria aerial labeling dataset. A composite training dataset for building segmentation was curated by combining images from all these opensource datasets. Multiple data sources from all over the world ensure that the resulting dataset can capture variability and diversity of building types, and building patterns, as well as the shapes, sizes, and colors of the buildings across space. All the images were then chipped down to $224 \times 224$ pixels. However, the chipping process resulted in a lot of instances containing either very few or no building pixels at all, which lowers the label density in the dataset. Such instances can directly affect the combined dataset by introducing a skew in the label distribution and causing model bias during training. To avoid bias and maintain balance in the combined dataset in terms of label density distribution, a High Label Filter (HLF) (\autoref{eq-hlf}) was defined as a ratio of building pixels to total image pixels for an image. An HLF threshold of 0.3 was determined by quantitative and observational analyses of the chipped image-label pairs across all datasets. This essentially means that only those image instances that have at least 30\% building pixels will be selected for the training process. Finally, a multiscale and multi-sensor dataset comprising 10,895 aerial and satellite image-building-label pairs was obtained.

\begin{equation}
    HLF = \frac{\Sigma_{i=0}^{224\times224}(p_i|p\in B)}{\Sigma_{i=0}^{224\times224}p_i}
    \label{eq-hlf}
\end{equation}
where $p$ represents every pixel of the image, and $B$ represents the set of building pixels of that image.

\subsection{Generating Guiding Features}

\begin{figure}
  \centering
  \includegraphics[width=01.0\linewidth]{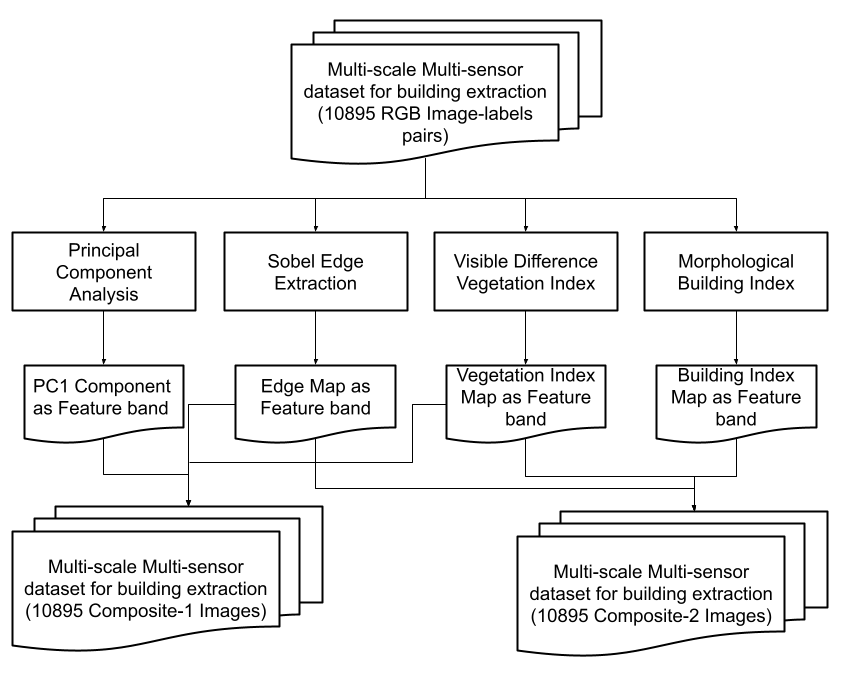}
  \caption{Methodology for extracting guiding features and generating different image composites for the dataset.}
  \label{fig:methodology-for-guiding-features}
\end{figure}

Deep learning gives comparable results and sometimes even better results when compared with benchmark methods for building segmentation, in the case of only RGB data \cite{Wang2020, Yang2018, Zuo2017}. From RGB data alone, a lot of spatial and contextual information can be extracted which can aid in better learning for building segmentation. This section discussed how subsidiary information extracted from RGB images can be used for guiding the learning process by incriminating more spatial context and providing more spectral discrimination to the features of interest which in this case are buildings. Feature extraction methods described below have been implemented on the curated multiscale multi-sensor dataset, and each piece of information is then used as a separate feature band to generate a composite image dataset. Hence, we constructed three multiscale datasets comprising the same 10,895 images but with different band combinations: Composite-0 (RGB), Composite-1(Sobel as Red, VDVI as Green, PC1 as Blue), and Composite-2 (Sobel as Red, VDVI as Green, MBI as Blue). We select these composites to have three bands to imitate the RGB setting, as done in previous studies \cite{Luo2021, Xu2018}. Moreover, the bands for Composite-1 and Composite-2 are chosen based on the subset that showed the highest variance. The proposed deep neural network is then trained on these three band combinations. \autoref{fig:methodology-for-guiding-features} shows a general methodology for guiding information extraction and composite data generation.

Principal Component Analysis (PCA) is a widely used dimensionality reduction and information extraction method which helps to reduce the dimensionality of the data while preserving the maximum amount of information in the data. For this study, PCA has been applied to RGB images, inspired by Xu et al.’s approach \cite{Xu2018}. This means that each band – R,G, and B were plotted in a 3D feature space, and based on their correlation with each other, they were orthogonally transformed to generate the principal components (PC). Each PC contains an Eigen Value which indicates the amount of information presenting in that PC. After applying PCA on all 10,895 RGB images, the first PC (PC1) containing the maximum information was extracted.

Edges are an important feature for accurate building segmentation as they create a definite bound for every probable building region. After qualitative and observational analysis of various edge extractors \cite{Xu2018, Zhao2018, Huang2012, Niveetha2012}, the Sobel operator was chosen to extract edges and generate edge-based feature bands. Edge detection filters are essentially high-pass filters that extract edges by up-shooting the first differential of the image contrast. The Sobel operator consists of two matrices, one of which is a 90\textdegree rotation of the other, for horizontal and vertical edge detection respectively (\autoref{eq-gx}, \autoref{eq-gy}). To highlight the edges, the magnitudes of convolution of both matrices are taken (\autoref{eq-g}). Since there is a dedicated matrix for horizontal and vertical edge detection, all edges lying between horizontal 0\textdegree and vertical 90\textdegree are effectively extracted.

\begin{equation}
    G_x = \begin{bmatrix}
-1 & 0 & 1\\
-2 & 0 & 2 \\
-1 & 0 & 1
\end{bmatrix}
\label{eq-gx}
\end{equation}

\begin{equation}
    G_y = \begin{bmatrix}
1 & 2 & 1\\
0 & 0 & 0 \\
-1 & -2 & -1
\end{bmatrix}
\label{eq-gy}
\end{equation}

\begin{equation}
    |G| = \sqrt{{G_x}^2 + {G_y}^{2}}
    \label{eq-g}
\end{equation}
where where $G_x$ is the vertical edge gradient, $G_y$ is the horizontal edge gradient, and $G$ is the resultant magnitude edge gradient 

Misclassification of buildings due to the spectral similarity between building roofs, lawns, and grass has been a prominent issue in building segmentation \cite{boonpook2018deep}. To increase the separability of buildings from vegetation class in general, vegetation indices have been widely used \cite{Liu2018semantic, Xu2018}. Given the dataset of this study, an RGB-based vegetation index called Visible Difference Vegetation Index (VDVI) (\autoref{eq-vdvi}) has been used \cite{Tan2018morphfilter, Wang2015vdviuav}. Its value lies between -1 and 1 like any other vegetation index. VDVI is applied on the multiscale RGB dataset to acquire the VDVI feature bands.

\begin{equation}
    VDVI = \frac{2 \times \rho_{green} - \rho_{red} - \rho_{blue}}{2 \times \rho_{green} + \rho_{red} + \rho_{blue}}
    \label{eq-vdvi}
\end{equation}
where $\rho_{green}, \rho_{red} and \rho_{blue}$ are the green, red and blue channels/bands of the image respectively.

While it is necessary to discriminate buildings with other spectrally similar classes, it is equally important to incriminate buildings against all other classes. For this purpose, Morphological Building Index (MBI) is used in this study. MBI is a technique directly dependent on the implicit and typical characteristics of buildings in a remote sensing image – brightness, size, and contrast \cite{Huang2012}. This is mathematically quantified by relating these characteristics with morphological operators such as top-hat reconstruction, granulometry, directionality, and spectral difference \cite{jiminez2017mbi}. The implementation of MBI consists of three main steps: Brightness calculation (\autoref{eq-b}), White top-hat reconstruction to obtain a morphological profile highlighting bright features (\autoref{eq-w}) and differentiating the profile (\autoref{eq-w'}) to obtain a monochrome MBI image ranging from 0 to 1. MBI was applied to the multiscale RGB dataset to acquire MBI feature bands.

\begin{equation}
    b(x) = \max_{1 \leq i \leq N} band_i(x)
    \label{eq-b}
\end{equation}

\begin{equation}
    W_{TopHat}(d,s) = b - \gamma_b^{re}(d,s)
    \label{eq-w}
\end{equation}

\begin{equation}
    W'_{TopHat}(d,s) = |W_{TopHat}(d, s + \Delta s) - W_{TopHat}(d,s)|
    \label{eq-w'}
\end{equation}
where $band_i(x)$ indicates spectral value of pixel $x$ at $i^{th}$ spectral band; $N$ indicates the number of bands; $d$ is the direction of linear building structural element; $s$ is the length of the linear building structural element; $b$ is the pixel brightness value; $\gamma_b^{re}()$ indicates the value of the brightest pixel after opening by reconstruction operation; $\Delta s$ indicates the interval of median lengths between maximum length of linear building structural element.

This resulted in four additional guiding feature bands namely PC1, Sobel, VDVI, and MBI apart from the three R, G, and B bands. These 7 bands together were then used to create 3 different band combinations (chosen exhaustively with the highest variance) namely: Composite-0, Composite-1, and Composite-2 (hereafter referred to as CB0, CB1, and CB2 respectively). CB0 is the original RGB band combination. CB1 comprises of Sobel band in red, the VDVI band in green, and the PC1 band in blue channels. CB2 comprises of Sobel band in red, the VDVI band in green, and the MBI band in blue channels. Figure 4 shows the RGB, PC1, Sobel, VDVI, MBI, Composite-1, and Composite-2 band combinations of three UAV images. These composite variations of data fusion serve as guiding features to the consequent dynamic Res-U-Net architecture to extract buildings, discussed in the next section.

\begin{figure}
  \centering
  \includegraphics[width=01.0\linewidth]{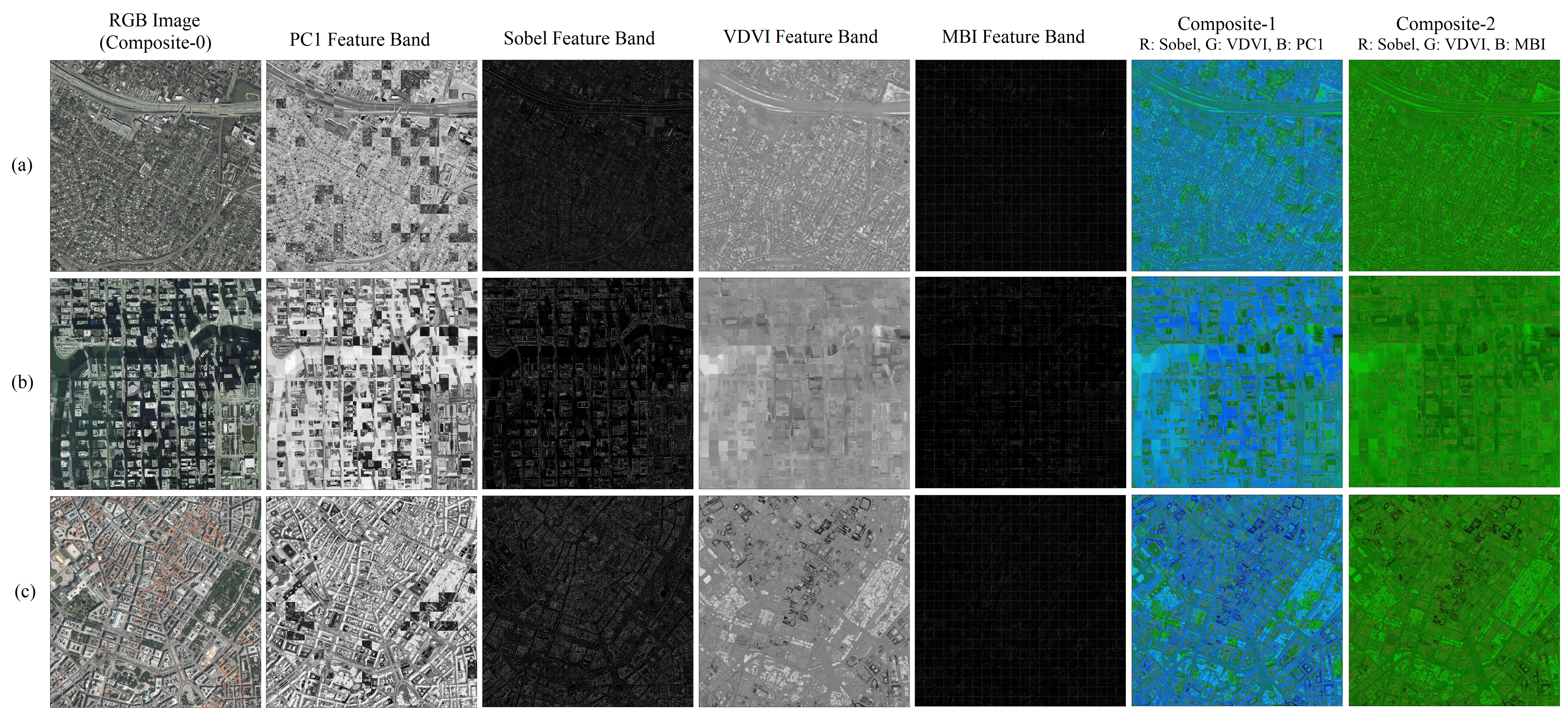}
  \caption{Three UAV images exhibiting the feature bands and Composite-0, Composite-1, and Composite-2 combinations. (a) Building class with road and open areas (b) Building class with shadow interference (c) Building class closely spaced and varying shapes}
  \label{fig:composite-samples}
\end{figure}

\subsection{Deep Neural Network - Architecture and Training}

In this study, we propose a dynamic Res-U-Net architecture to learn multiscale building segmentation from RGB imagery. Res-U-Net is a combination of two deep learning architectures: 1) U-Net, a Fully Convolutional Network (FCN) with an encoder-decoder architecture, frequently used for building segmentation \cite{Erdem2020}; and 2) ResNet34, a 34-layer deep convolutional neural network which is also a state-of-the-art model for image classification. The dynamic architecture is inspired by the implementation in FastAI, a python package, that initializes the decoder of a UNet dynamically, and automatically infers the sizes of intermediate blocks based on the initial dimension of the corresponding encoder \cite{Zhuang2019}. Each encoder-decoder block of the proposed Res-U-Net comprises a series of 2D convolution layers, batch normalization, and ReLU activations. Res-U-Net learns iteratively from edges to shapes, and then to regions, and objects. In this case, it will grow a building region from a single seed pixel, restrict the region with a shape boundary and then validate those groups of pixels as a single building feature.

To enable multiscale learning within the proposed Res-U-Net architecture, each layer is initialized at an upscale factor of 2 between the range of 0.4 to 3 per the spatial resolutions of the multiscale dataset. There have been studies that use attention pyramid networks \cite{Tian2022}, Siamese networks \cite{He2019} as well as transformer-based architectures \cite{chen2022multiscale} to attempt multiscale building segmentation. However, this study uses PixelShuffle \cite{Shi2016subpixel, Shi2016subpixel}, a deconvolution-based approach to simulate super-resolution-like phenomena, to enable multiscale learning. The advantage of using PixelShuffle is that it is adaptive to multiple stride sizes, and hence can be suffixed to any encoder-decoder-based architecture. Towards the end of the network, multiple pixelshuffle blocks are created to imitate the effect of de-convolution. While PixelShuffle has been a popular method used for super-resolution \cite{Noh2015}, this study uses it for multiscale building segmentation. \autoref{tbl-network_architecture} describes the layer-wise architecture of the proposed Res-U-Net.

\begin{table}
\caption{Layer-wise architecture specifications of the Res-U-Net for building segmentation}
\label{tbl-network_architecture}
\begin{tabular}{|lll|l|l|l|}
\hline
\multicolumn{3}{|l|}{Layer}                                                                              & Kernel Size & Output Shape    & Stride \\ \hline
\multicolumn{1}{|l|}{Conv2d}                                & \multicolumn{1}{l|}{}       &              & 7 x 7       & 64 x 112 x 112  & 2      \\ \hline
\multicolumn{1}{|l|}{\multirow{3}{*}{Sequential Block 1}}   & \multicolumn{1}{l|}{Conv2d} &              & 3 x 3       & 64 x 56 x 56    & 2      \\ \cline{2-6} 
\multicolumn{1}{|l|}{}                                      & \multicolumn{1}{l|}{Conv2d} &              & 3 x 3       & 64 x 56 x 56    & 1      \\ \cline{2-6} 
\multicolumn{1}{|l|}{}                                      & \multicolumn{1}{l|}{Conv2d} &              & 3 x 3       & 64 x 56 x 56    & 1      \\ \hline
\multicolumn{1}{|l|}{\multirow{5}{*}{Sequential Block 2}}   & \multicolumn{1}{l|}{}       & Down Block 1 & 1 x 1       & 128 x 28 x 28   & 2      \\ \cline{2-6} 
\multicolumn{1}{|l|}{}                                      & \multicolumn{1}{l|}{Conv2d} &              & 3 x 3       & 128 x 28 x 28   & 1      \\ \cline{2-6} 
\multicolumn{1}{|l|}{}                                      & \multicolumn{1}{l|}{Conv2d} &              & 3 x 3       & 128 x 28 x 28   & 1      \\ \cline{2-6} 
\multicolumn{1}{|l|}{}                                      & \multicolumn{1}{l|}{Conv2d} &              & 3 x 3       & 128 x 28 x 28   & 1      \\ \cline{2-6} 
\multicolumn{1}{|l|}{}                                      & \multicolumn{1}{l|}{Conv2d} &              & 3 x 3       & 128 x 28 x 28   & 1      \\ \hline
\multicolumn{1}{|l|}{\multirow{6}{*}{Sequential Block 3}}   & \multicolumn{1}{l|}{}       & Down Block 2 & 1 x 1       & 256 x 14 x 14   & 2      \\ \cline{2-6} 
\multicolumn{1}{|l|}{}                                      & \multicolumn{1}{l|}{Conv2d} &              & 3 x 3       & 256 x 14 x 14   & 1      \\ \cline{2-6} 
\multicolumn{1}{|l|}{}                                      & \multicolumn{1}{l|}{Conv2d} &              & 3 x 3       & 256 x 14 x 14   & 1      \\ \cline{2-6} 
\multicolumn{1}{|l|}{}                                      & \multicolumn{1}{l|}{Conv2d} &              & 3 x 3       & 256 x 14 x 14   & 1      \\ \cline{2-6} 
\multicolumn{1}{|l|}{}                                      & \multicolumn{1}{l|}{Conv2d} &              & 3 x 3       & 256 x 14 x 14   & 1      \\ \cline{2-6} 
\multicolumn{1}{|l|}{}                                      & \multicolumn{1}{l|}{Conv2d} &              & 3 x 3       & 256 x 14 x 14   & 1      \\ \hline
\multicolumn{1}{|l|}{\multirow{6}{*}{Sequential Block 4}}   & \multicolumn{1}{l|}{}       & Down Block 3 & 1 x 1       & 512 x 7 x 7     & 2      \\ \cline{2-6} 
\multicolumn{1}{|l|}{}                                      & \multicolumn{1}{l|}{Conv2d} &              & 3 x 3       & 512 x 7 x 7     & 1      \\ \cline{2-6} 
\multicolumn{1}{|l|}{}                                      & \multicolumn{1}{l|}{Conv2d} &              & 3 x 3       & 512 x 7 x 7     & 1      \\ \cline{2-6} 
\multicolumn{1}{|l|}{}                                      & \multicolumn{1}{l|}{Conv2d} &              & 3 x 3       & 512 x 7 x 7     & 1      \\ \cline{2-6} 
\multicolumn{1}{|l|}{}                                      & \multicolumn{1}{l|}{Conv2d} &              & 3 x 3       & 512 x 7 x 7     & 1      \\ \cline{2-6} 
\multicolumn{1}{|l|}{}                                      & \multicolumn{1}{l|}{Conv2d} &              & 3 x 3       & 1024 x 7 x 7    & 1      \\ \hline
\multicolumn{1}{|l|}{\multirow{3}{*}{U-Net Block 1}}        & \multicolumn{1}{l|}{}       & pixelshuffle & -           & 256 x 14 x 14   & -      \\ \cline{2-6} 
\multicolumn{1}{|l|}{}                                      & \multicolumn{1}{l|}{Conv2d} &              & 3 x 3       & 512 x 14 x 14   & 1      \\ \cline{2-6} 
\multicolumn{1}{|l|}{}                                      & \multicolumn{1}{l|}{Conv2d} &              & 3 x 3       & 512 x 14 x 14   & 1      \\ \hline
\multicolumn{1}{|l|}{Connector}                             & \multicolumn{1}{l|}{Conv2d} &              & 3 x 3       & 1024 x 14 x 14  & 1      \\ \hline
\multicolumn{1}{|l|}{\multirow{3}{*}{U-Net Block 2}}        & \multicolumn{1}{l|}{}       & pixelshuffle & -           & 256 x 28 x 28   & -      \\ \cline{2-6} 
\multicolumn{1}{|l|}{}                                      & \multicolumn{1}{l|}{Conv2d} &              & 3 x 3       & 512 x 14 x 14   & 1      \\ \cline{2-6} 
\multicolumn{1}{|l|}{}                                      & \multicolumn{1}{l|}{Conv2d} &              & 3 x 3       & 512 x 14 x 14   & 1      \\ \hline
\multicolumn{1}{|l|}{Connector}                             & \multicolumn{1}{l|}{Conv2d} &              & 3 x 3       & 1024 x 14 x 14  & 1      \\ \hline
\multicolumn{1}{|l|}{\multirow{3}{*}{U-Net Block 3}}        & \multicolumn{1}{l|}{}       & pixelshuffle & -           & 192 x 56 x 56   & -      \\ \cline{2-6} 
\multicolumn{1}{|l|}{}                                      & \multicolumn{1}{l|}{Conv2d} &              & 3 x 3       & 256 x 56 x 56   & 1      \\ \cline{2-6} 
\multicolumn{1}{|l|}{}                                      & \multicolumn{1}{l|}{Conv2d} &              & 3 x 3       & 256 x 56 x 56   & 1      \\ \hline
\multicolumn{1}{|l|}{Connector}                             & \multicolumn{1}{l|}{Conv2d} &              & 3 x 3       & 512 x 56 x 56   & 1      \\ \hline
\multicolumn{1}{|l|}{\multirow{3}{*}{U-Net Block 4}}        & \multicolumn{1}{l|}{}       & pixelshuffle & -           & 128 x 112 x 112 & 1      \\ \cline{2-6} 
\multicolumn{1}{|l|}{}                                      & \multicolumn{1}{l|}{Conv2d} &              & 3 x 3       & 96 x 112 x 112  & 1      \\ \cline{2-6} 
\multicolumn{1}{|l|}{}                                      & \multicolumn{1}{l|}{Conv2d} &              & 3 x 3       & 96 x 112 x 112  & 1      \\ \hline
\multicolumn{1}{|l|}{\multirow{2}{*}{PixelShuffle\_ICNR}}   & \multicolumn{1}{l|}{Conv2d} &              & 1 x 1       & 348 x 112 x 112 & 1      \\ \cline{2-6} 
\multicolumn{1}{|l|}{}                                      & \multicolumn{1}{l|}{}       & pixelshuffle & -           & 96 x 224 x 224  & -      \\ \hline
\multicolumn{1}{|l|}{\multirow{2}{*}{Sequential Extension}} & \multicolumn{1}{l|}{Conv2d} &              & 3 x 3       & 99 x 224 x 224  & 1      \\ \cline{2-6} 
\multicolumn{1}{|l|}{}                                      & \multicolumn{1}{l|}{Conv2d} &              & 3 x 3       & 99 x 224 x 224  & 1      \\ \hline
\multicolumn{1}{|l|}{Pooler}                                & \multicolumn{1}{l|}{Conv2d} &              & 1 x 1       & 2 x 224 x 224   & 1      \\ \hline
\end{tabular}
\end{table}

To simultaneously preserve the boundary and shape of building segmentation, a unique combination of Binary Cross Entropy (BCE) loss (\autoref{eq-bceloss}) and dice loss (\autoref{eq-dice}) is used to train the proposed Res-U-Net. BCE loss is a probability distribution-based loss \cite{Zhang2018bceloss} used to minimize the entropy between the model prediction and the ground truth. It also preserves the crispness near the boundary regions. Dice loss is a region-based Intersection-over-Union-like metric \cite{Sudre2017} and that maximizes the overlap and similarity between the predicted region and the ground truth of the building region in the image. BCE loss and Dice loss were combined using Combo Loss (\autoref{eq-comboloss}) which focused on both boundary and region preservation. To control the contribution of Dice loss in Combo Loss, $\alpha$ was used as a hyperparameter, and for this study, $\alpha=1$ was used to tune the Combo Loss.

\begin{equation}
    BCE_{Loss} = -\frac{1}{ps} \Sigma_{i=1}^{ps}g_i \times \log(p_i) + (1 - g_i) \times \log(1 - p_i)
    \label{eq-bceloss}
\end{equation}

\begin{equation}
    Dice_{Loss} = \frac{2 \times \Sigma_{i=0}^{ps}p_ig_i}{\Sigma_{i=0}^{ps}{p_i}^2 + \Sigma_{i=0}^{ps}{g_i}^2}
    \label{eq-dice}
\end{equation}

\begin{equation}
    ComboLoss = BCE_{Loss} + \alpha \times Dice_{Loss}
    \label{eq-comboloss}
\end{equation}
where $g$ is the ground truth/pixel label (1 if building, 0 if non-building); $p$ is the probability of that pixel being a building pixel (model output); and $ps$ is patch size of image tiles (224 x 224 in this case).

\begin{figure}
  \centering
  \includegraphics[width=01.0\linewidth]{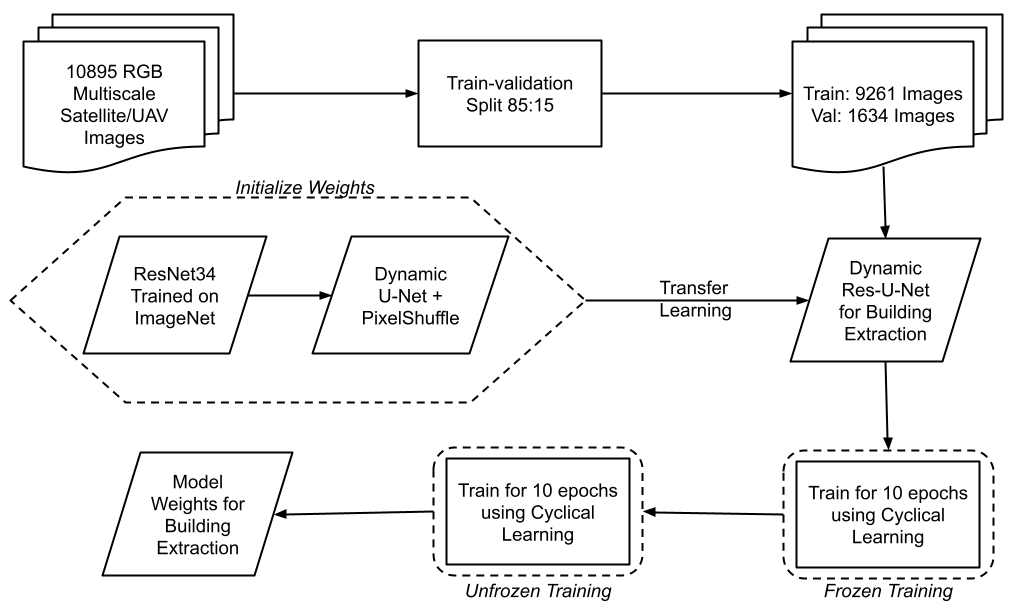}
  \caption{Three UAV images exhibiting the feature bands and Composite-0, Composite-1, and Composite-2 combinations. (a) Building class with road and open areas (b) Building class with shadow interference (c) Building class closely spaced and varying shapes}
  \label{fig:overall-methodology}
\end{figure}

In this study, an attempt was made to optimize the resource and time consumption in training the network without decreasing the learning parameters while preserving the quality of results. This was done by employing concepts of Cyclical Learning Rates (CLR) and SuperConvergence (SC), by harmonically oscillating the learning rate within a fixed range as the training progresses \cite{Smith2019superconvergence, Smith2017cyclicalLR}. Apart from this, an unconventional training methodology that alters the order in which the layers of the network are trained, is used for quicker convergence. \autoref{fig:overall-methodology} describes the step-by-step training methodology. The network is trained separately for CB0, CB1, and CB2 band combinations. Each combination contained 10,895 3-band image-label pairs. Train and validation instances are split at a ratio of 85:15, resulting in 9261 instances for training and the remaining 1634 for validation. The three-channel encoder of the Dynamic U-Net was initialized with the weights of RGB channels of the pre-trained ResNet34, forming the dynamic Res-U-Net. Transfer learning was used to initialize Res-U-Net, since ResNet34 is state of the art for image classification, and building segmentation is a synonymous problem. After initializing the Res-U-Net, all layers of the network except the last layer are frozen and the network is trained for 15 epochs using cyclical learning. Such kind of practice is commonly used in transfer learning to optimize training performance \cite{gulli2017deep}. After frozen training, the network is unfrozen so that weights of all layers can be updated and are trained for another 15 epochs, both within a cyclical learning range of [0.0001, 0.001]. The network is optimized with ADAM optimizer at a decay rate of 0.9. A fallback provision was made during the training process to allow the model to fall back to its previous state if the consecutive epoch did not result in a better version of the model.

\section{Results}

In this section, we evaluate the model performance on the validation data and test sets using quantitative metrics. The proposed dynamic Res-U-Net architecture is evaluated on all three band combinations (CB0, CB1, and CB2) of images from both validation and test sets, and a comprehensive discussion for each of the three-band combinations follows. To test the performance of the proposed network for building segmentation, the model weights obtained after training are used to make predictions on the validation data, and a pixel-to-pixel comparison is made between the predicted output and the ground truth. Model predictions and ground truth are both in the form of binary images, with pixel having value ‘1’ in case of a building and ‘0’ otherwise. True Positive (TP), True Negative (TN), False Positive (FP) and False Negative (FN) counts are used to create a confusion map of the input image to specifically highlight which land cover areas were vulnerable to false counts and in which cases the model is making robust predictions. To quantify the prediction made by the model in terms of binary segmentation, the metrics of accuracy (\autoref{eq-accuracy}), precision (\autoref{eq-precision}), recall (\autoref{eq-recall}), and F1-score (\autoref{eq-f1}) were used. To further perform a feature-based evaluation, object-based metrics such as branching factor (\autoref{eq-bf}), miss factor (\autoref{eq-mf}), and Intersection over Union (IoU) (\autoref{eq-iou}) are used.

\begin{equation}
    accuracy = \frac{tp + tn}{tp + tn + fp + fn}
    \label{eq-accuracy}
\end{equation}

\begin{equation}
    precision = \frac{tp}{tp + fp}
    \label{eq-precision}
\end{equation}

\begin{equation}
    recall = \frac{tp}{tp + fn}
    \label{eq-recall}
\end{equation}

\begin{equation}
    F1 = 2 \times \frac{precision \times recall}{precision + recall}
    \label{eq-f1}
\end{equation}

\begin{equation}
    branchingFactor=\frac{fp}{tp}
    \label{eq-bf}
\end{equation}

\begin{equation}
    missFactor = \frac{fn}{tp}
    \label{eq-mf}
\end{equation}

\begin{equation}
    IoU = \frac{tp}{tp + fn + fp} \times 100\%
    \label{eq-iou}
\end{equation}
where $tp$ (True positive) is the number of pixels correctly predicted as buildings; $tn$ (True negative) is the number of pixels correctly predicted as non buildings; $fp$ (False positive) is the number of pixels incorrectly predicted as buildings; $fn$ is the number of building pixels incorrectly predicted as non buildings.

\subsection{In-sample Multiscale Validation on UAV Images}

\begin{table}
\caption{Building segmentation evaluation metrics of the proposed Res-U-Net for all three composites (CB0, CB1, CB2) of the validation set}
\label{tbl-insample_validation_results}
\begin{tabular}{|l|l|l|l|}
\hline
                      & CB0 (R,G,B) & CB1 (Sobel, VDVI, PC1) & CB2 (Sobel, VDVI, MBI) \\ \hline
Mean Accuracy         & 0.956       & 0.912                  & 0.932                  \\ \hline
Mean Precision        & 0.883       & 0.780                  & 0.80                   \\ \hline
Mean Recall           & 0.861       & 0.733                  & 0.722                  \\ \hline
Mean F1-Score         & 0.860       & 0.708                  & 0.756                  \\ \hline
Mean Branching Factor & 0.193       & 0.347                  & 0.287                  \\ \hline
Mean Miss Factor      & 0.230       & 0.559                  & 0.458                  \\ \hline
Mean IoU              & 0.80        & 0.645                  & 0.692                  \\ \hline
\end{tabular}
\end{table}

In-sample validation is performed to tune the model hyperparameters for all three band combinations. Table 2 shows a comparison of model performance across all three composites of the validation set. These are average metrics for all instances of the validation set. As mentioned earlier, CB0 is the original RGB image. CB1 and CB2 comprise secondary features derived from the RGB image (CB1: Sobel as Red, VDVI as Green, PC1 as Blue; CB2: Sobel as Red, VDVI as Green, MBI as Blue) to imitate the commonly used RGB image setting for this kind of segmentation problems \cite{Luo2021, Xu2018}. Out of CB0, CB1, and CB2, CB0 shows the highest accuracy and IoU (\autoref{tbl-insample_validation_results}), which indicates that the raw RGB feature combination is most advantageous to the proposed Res-U-Net. The accuracies of CB1 and CB2 are in a comparable ballpark, suggesting a potential opportunity of using PC1, VDVI, Sobel, and MBI as features to extract buildings more effectively where vegetation or shadow is more dominant in the background.

\subsection{Out-sample Multiscale Testing on UAV Images}

Out-sample testing was performed to assess the model performance all three combinations. The unseen UAV test set comprised six UAV images from across five different US and Austrian cities. The test images are a mix in terms of land-cover classes present in the image, building placement, and urban density and shadow distribution, creating a high-variance building segmentation test set. Figure 6 shows the results for building segmentation for UAV test images composites 0, 1, and 2. The first column is the input to the model, the second column is the ground truth, and then the predictions, as well as evaluation maps for each composite, follow. In the prediction map, a pixel is predicted as a building as shown in white and others in black. In the evaluation map, True Positives (TP) are in white, True Negatives (TN) in black, False Positives (FP) in red and False Negatives (FN) in yellow.

\begin{figure}
  \centering
  \includegraphics[width=01.0\linewidth]{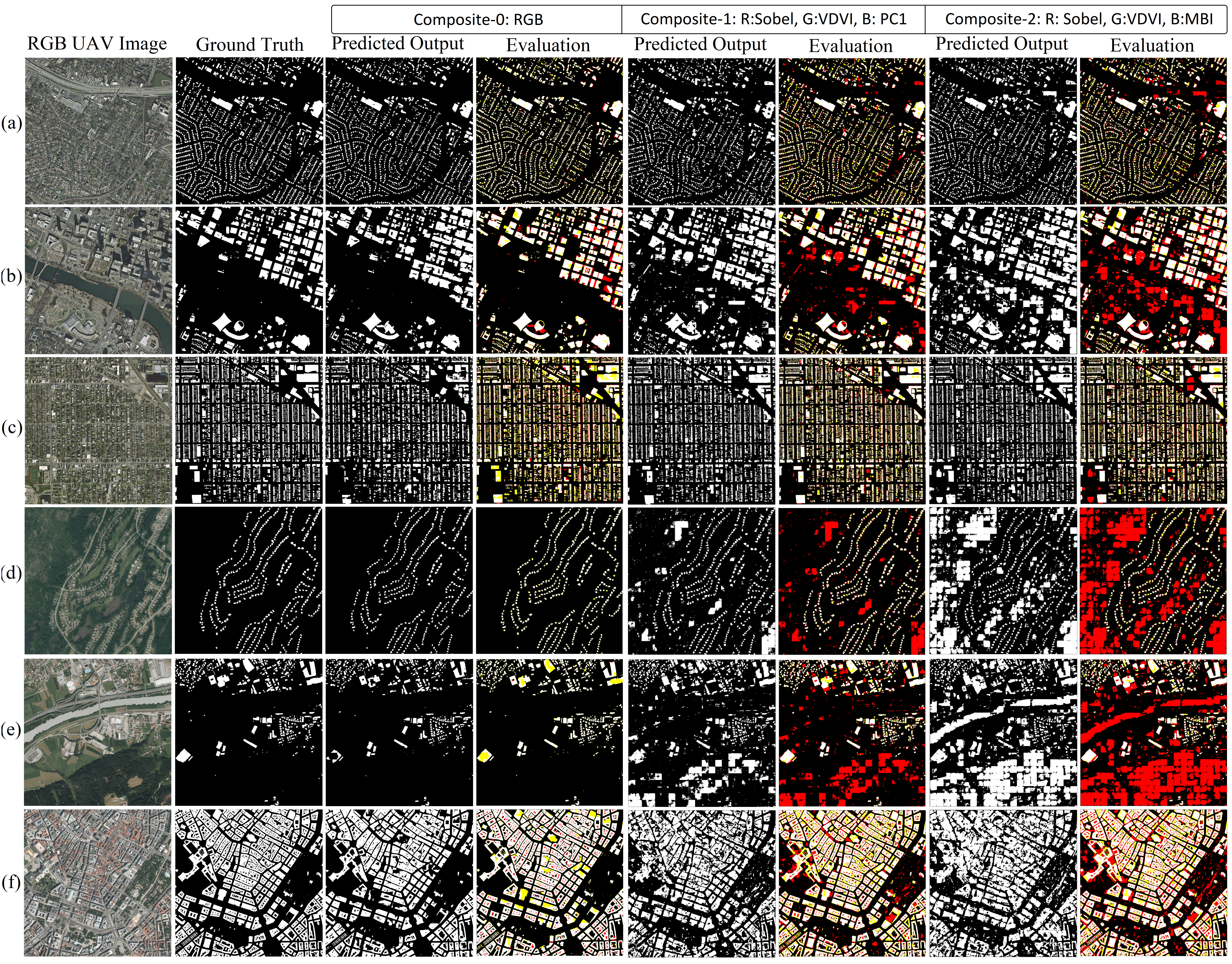}
  \caption{UAV test set instances and their building segmentation results for each of the three composites. The first column shows inputs to the model, the second column is the ground truth labels. The predictions as well as the evaluation maps of each band composition are shown in the last six columns. The evaluation maps show TP (white), TN (black), FP (red), and FN (yellow). (a), (b) From Austin, USA (c) From Chicago, USA (d) From Kitsap County, USA (e) From Tyrol West, Austria (f) From Vienna Austria}
  \label{fig:outsample-uav-testing}
\end{figure}

For CB0, \autoref{fig:outsample-uav-testing}(a), (c), and (f) are dense urban areas having closely spaced buildings, wherein the model is able to segment buildings with minimal false positives and successful instance segmentation. \autoref{fig:outsample-uav-testing}(b) is an urban area with high-rise buildings and significant shadowy regions. The proposed model does not misidentify shadowy areas as buildings, surpassing a prominent building segmentation challenge \cite{Huang2012}. However, the model does miss the buildings that are under shadows. \autoref{fig:outsample-uav-testing}(a), (b), and (f) show successful building segmentation in presence of spectrally similar features such as cemented roads and parking lots as well as spatially similar features such as roads, open grounds, and vegetation patches shaped like buildings. Figures 6 (d) and (e) contain a large cover of vegetation, and \autoref{fig:outsample-uav-testing}(b) and (e) contain a large area of water – which are spectrally conflicting landcover for the urban class, wherein the model segments the buildings successfully with high accuracy and low miss factors. \autoref{tbl-outsample_cb2_validation_results} shows the evaluation metrics for individual CB0 images in \autoref{fig:outsample-uav-testing}.

\begin{table}
\caption{Mean Evaluation metrics for the CB2 (Sobel, VDVI, MBI) UAV test set instances as shown in Figure 6}
\label{tbl-outsample_cb2_validation_results}
\begin{tabular}{|l|l|l|l|l|l|l|l|}
\hline
            & Accuracy & Precision & Recall & F1-Score & Branching Factor & Miss Factor & IoU \\ \hline
\autoref{fig:outsample-uav-testing}(a) & 0.917         & 0.753          & 0.699       & 0.725         & 0.328                 & 0.431            & 0.568    \\ \hline
\autoref{fig:outsample-uav-testing}(b) & 0.870         & 0.549          & 0.861       & 0.671         & 0.821                 & 0.162            & 0.504    \\ \hline
\autoref{fig:outsample-uav-testing}(c) & 0.888         & 0.830          & 0.757       & 0.792         & 0.204                 & 0.321            & 0.655    \\ \hline
\autoref{fig:outsample-uav-testing}(d) & 0.753         & 0.165          & 0.750       & 0.270         & 5.062                 & 0.334            & 0.156    \\ \hline
\autoref{fig:outsample-uav-testing}(e) & 0.694         & 0.196          & 0.826       & 0.317         & 4.099                 & 0.211            & 0.188    \\ \hline
\autoref{fig:outsample-uav-testing}(f) & 0.768         & 0.717          & 0.782       & 0.748         & 0.395                 & 0.279            & 0.597    \\ \hline
\end{tabular}
\end{table}

\subsection{Out-sample Uni-scale Test on Satellite Imagery}

\begin{figure}
  \centering
  \includegraphics[width=0.85\linewidth]{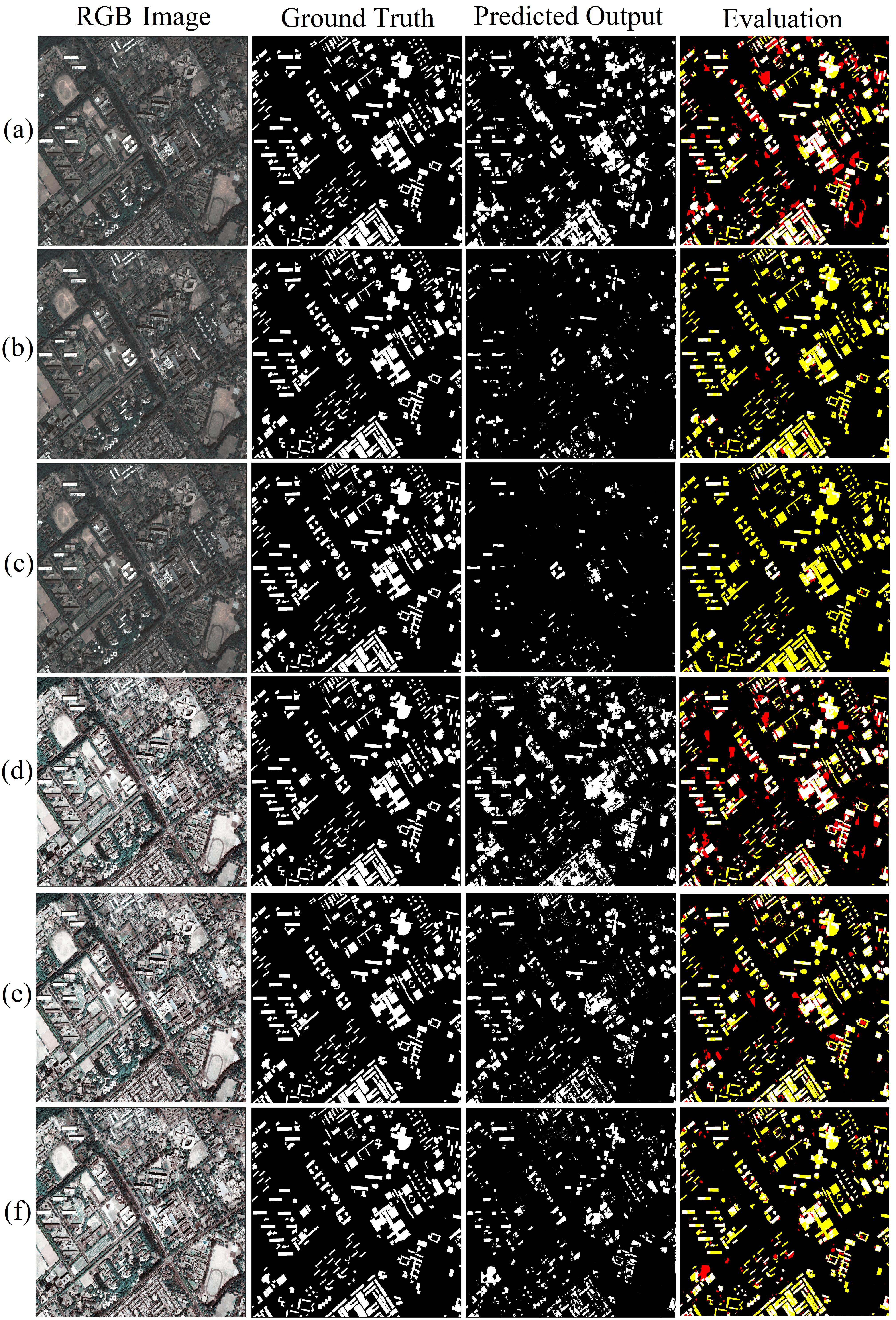}
  \caption{Satellite test set evaluation. The first column is input to the model, the second column is the ground truth, the third column shows model predictions for building segmentation, and the fourth column is evaluation images showing TP (white), TN (black), FP (red), and FN (yellow). (a) CB0 un-equalized (b) CB1 un-equalized (c) CB2 un-equalized (d) CB0 equalized (e) CB1 equalized (f) CB2 equalized}
  \label{fig:outsample-satellite-testing}
\end{figure}

For the out-sample uni-scale testing, a scene of the Chandigarh city area (shown in the study area section) captured on Oct 27, 2016, from the WorldView-3 satellite sensor was obtained from the Indian Institute of Remote Sensing (IIRS), Indian Space Research Organization (ISRO). Although the WorldView-3 sensor has 8 spectral bands from 0.31m to 1.4m spatial resolutions, only the RGB channels at 1.24m are chosen for the testing image with dimensions as $3136 \times 2688$ pixels. For testing, the satellite image is contrast-corrected using the histogram equalization method. For both raw and contrast equalized images, the proposed model is tested on CB0, CB1, and CB2 band combinations. \autoref{fig:outsample-satellite-testing} shows the building segmentation results for raw RGB images as well as its feature extracted band combinations (\autoref{fig:outsample-satellite-testing} (a)-(c)) and for contrast equalized images (\autoref{fig:outsample-satellite-testing} (d)-(f)). The first column is the input to the model, the second column is the ground truth, the third column is the segmented building map as predicted by the model and the fourth column shows the evaluation of the prediction with True Positives (TP) in white, True Negatives (TN) in black, False Positives (FP) in red and False Negatives (FN) in yellow. A general observation noted across all composites is that because of contrast equalization, there is an overestimation in building segmentation with high true and false positives as compared to building segmentation from the un-equalized image. \autoref{tbl-outsample_satellite_validation_results} shows the evaluation metrics for all instances of \autoref{fig:outsample-satellite-testing}.

Model performance on un-equalized CB0 (\autoref{fig:outsample-satellite-testing}(a)) and equalized CB0 (\autoref{fig:outsample-satellite-testing}(d)) is favorable. The model is able to extract buildings of all shapes and curves and performs well even in vegetative surroundings. Typical challenges are faced in the case of open playgrounds and sandy grounds. Contrast equalization increases true positives as well as false positives. Hence, while the model detects more buildings than it did in the un-equalized image, it also falsely detected more non-building pixels as buildings.

For CB1, model performances on both un-equalized (\autoref{fig:outsample-satellite-testing}(b)) and equalized (\autoref{fig:outsample-satellite-testing}(e)) are moderate to fair as the model could detect only around half the buildings present in the image. As compared to CB0, the false positive rate grossly increases in CB1 (\autoref{fig:outsample-satellite-testing}(b)), potentially because the principal components contain more spatial information and lose spectral information after PCA. Equalizing CB1 image significantly improves the results, and the model is able to extract more buildings, though still with a slightly higher false positive rate.

Model performances on un-equalized CB2 (\autoref{fig:outsample-satellite-testing}(b)) and equalized composite-1 (\autoref{fig:outsample-satellite-testing}(e)) are moderate as the model is able to detect even lesser buildings as compared to CB1. This could be because of the MBI feature band which creates discrimination between buildings and shadows by accounting for luminance. As a result, all buildings reflecting strongly and brightly are extracted accurately while the buildings slightly under shadows or the ones which are not reflecting so brightly were not extracted at all. Equalization improves the model performance on composite-2 by decreasing the false negative rate significantly, as can be noted from the metrics in \autoref{tbl-outsample_satellite_validation_results}.

\begin{table}
\caption{Evaluation metrics for Satellite test set instances as shown in Figure 7}
\label{tbl-outsample_satellite_validation_results}
\begin{tabular}{|l|l|l|l|l|l|l|l|}
\hline
            & Accuracy & Precision & Recall & F1-Score & Branching Factor & Miss Factor & IoU   \\ \hline
CB0 un-eq \autoref{fig:outsample-satellite-testing}(a) & 0.891    & 0.687     & 0.664  & 0.675    & 0.506            & 0.505       & 0.497 \\ \hline
CB0 eq \autoref{fig:outsample-satellite-testing}(d) & 0.90     & 0.663     & 0.701  & 0.681    & 0.455            & 0.426       & 0.531 \\ \hline
CB1 un-eq \autoref{fig:outsample-satellite-testing}(b) & 0.873    & 0.893     & 0.245  & 0.385    & 0.119            & 3.069       & 0.239 \\ \hline
CB1 eq \autoref{fig:outsample-satellite-testing}(e) & 0.902    & 0.807     & 0.518  & 0.631    & 0.239            & 0.930       & 0.461 \\ \hline
CB2 un-eq \autoref{fig:outsample-satellite-testing}(c) & 0.862    & 0.915     & 0.158  & 0.270    & 0.092            & 5.318       & 0.156 \\ \hline
CB2 eq \autoref{fig:outsample-satellite-testing}(f) & 0.895    & 0.818     & 0.447  & 0.578    & 0.222            & 1.235       & 0.407 \\ \hline
\end{tabular}
\end{table}

\section{Discussion}

Apart from proposing a Res-U-Net that can segment buildings from multiple scales and proposing a combined loss function, this study also implemented optimization policies for training a deep network. This section discusses the impact of the Combo Loss on building segmentation, as well as the advantage of using CLR and SC for training the network in terms of time and resource optimization. We also discuss the challenges faced in this study for building segmentation in some typical scenarios. Finally, we conclude with a comparison of the proposed method with benchmark building segmentation methods.

\subsection{Impact of Combo-Loss on Spatial Contextual Learning}

\begin{figure}
  \centering
  \includegraphics[width=0.85\linewidth]{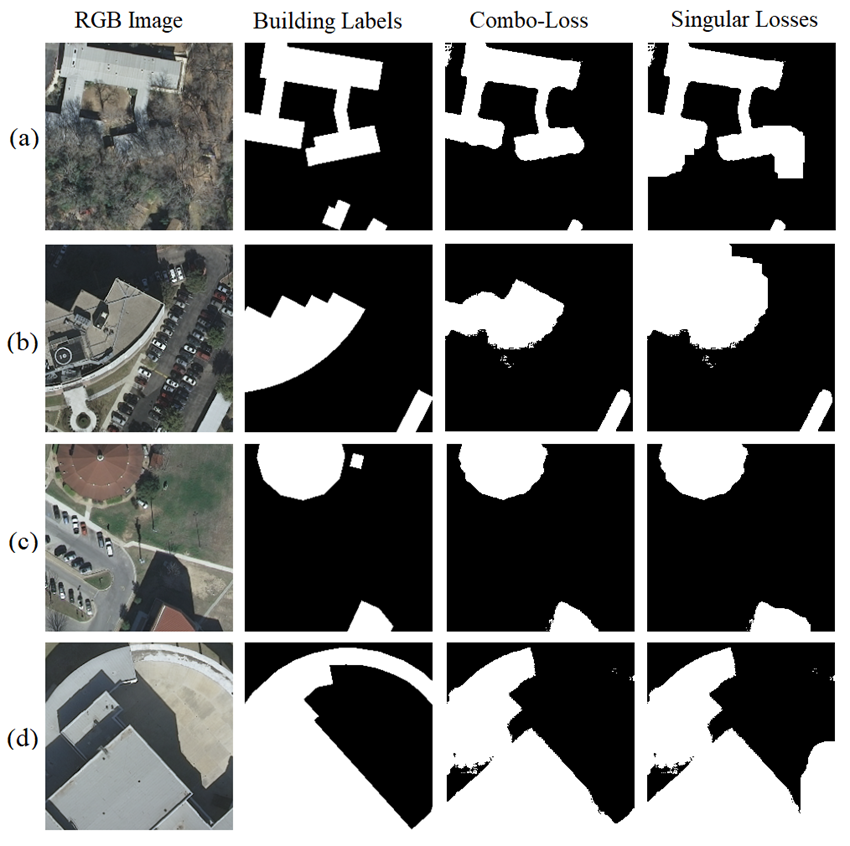}
  \caption{Comparison of Combo-Loss and singular loss for building segmentation. The first column contains the RGB Image, the second column contains corresponding image labels, the third column contains building segmentation results when Combo-Loss is used, the fourth column contains building segmentation results when other singular loss functions are used; in case of (a) surrounding vegetation (b) irregular building shapes and surrounding shadows (c) irregular building shapes in spectrally similar classes (d) irregular building shapes and varying undulation}
  \label{fig:loss-comparison}
\end{figure}

A novel combination of Binary Cross Entropy (BCE) Loss and Dice Loss called the Combo-Loss (\autoref{eq-comboloss}) was used to train the network. The motivation for using such a Combo-Loss was that it preserved the spatial contextual information very effectively. Being a combination of BCE and Dice, it accounts for both boundary and edge structure as well as the regional extent of a building object spread across multiple pixels. This was particularly helpful in extracting buildings having atypical and irregular shapes as well as blurred boundaries. An infamous problem in building segmentation in a spectrally restricted dataset has always been the leakage of predicted segments outside the building edge in the case of an irregularly shaped building \cite{Liu2018semantic, He2016}. Moreover, a lot of times, surrounding trees or vegetation as well as non-building classes under shadows result in significant false positives \cite{Xu2018, Huang2012, Huang2011}. Using the Combo-Loss led to combined advantages of both the loss functions. The BCE component improved the crispness of extracted building edges and adapted the model to discriminate against all non-building classes, even those under shadows, hence decreasing false positives due to surrounding vegetation and shadows. The Dice component mainly improved building segmentation for irregular shapes and sizes. \autoref{fig:loss-comparison} shows the comparison of proposed Combo-Loss and other singular loss functions such as BCE loss or Dice loss, for building segmentation for select instances of surrounding vegetation (\autoref{fig:loss-comparison}(a)), surrounding spectrally similar classes (\autoref{fig:loss-comparison}(d)) and surrounding shadows (\autoref{fig:loss-comparison}(b,c,d)); wherein the distinction between all three loss functions is visually evident.

The superior performance of Combo-Loss was also confirmed by quantitative analysis. Three separate versions of the same Res-U-Net were trained for the CB0 combination, one for each of the Combo-Loss, Dice Loss and BCE Loss. The image patches in Figure-8 were then inferred using all three models and IoU was calculated for each combimation instance. \autoref{tbl-loss_comparison} shows the IoU for each instance of \autoref{fig:loss-comparison}, for Combo-Loss and the higher value among Dice Loss or BCE Loss. It can be noted that the Combo-Loss reported a consistently higher IoU in all cases. Visually (and qualitatively), it can be observed from \autoref{fig:loss-comparison}, that using an individual loss function over the Combo-Loss always resulted in an ‘overflow’ of the prediction onto a non-building area.

\begin{table}[h]
\caption{Quantitative comparison between Combo Loss, BCE Loss, and Dice Loss using the IoU metric}
\label{tbl-loss_comparison}
\begin{tabular}{|l|l|l|}
\hline
                                 & $ComboLoss_{IoU}$ & $\max(DiceLoss_{IoU}, BCELoss_{IoU})$ \\ \hline
\autoref{fig:loss-comparison}(a) & 0.66           & 0.59                             \\ \hline
\autoref{fig:loss-comparison}(b) & 0.53           & 0.48                             \\ \hline
\autoref{fig:loss-comparison}(c) & 0.76           & 0.61                             \\ \hline
\autoref{fig:loss-comparison}(d) & 0.43           & 0.27                             \\ \hline
\end{tabular}
\end{table}

\subsection{Impact of Using Cyclical Learning and SuperConvergence}

\begin{figure}
  \centering
  \includegraphics[width=00.85\linewidth]{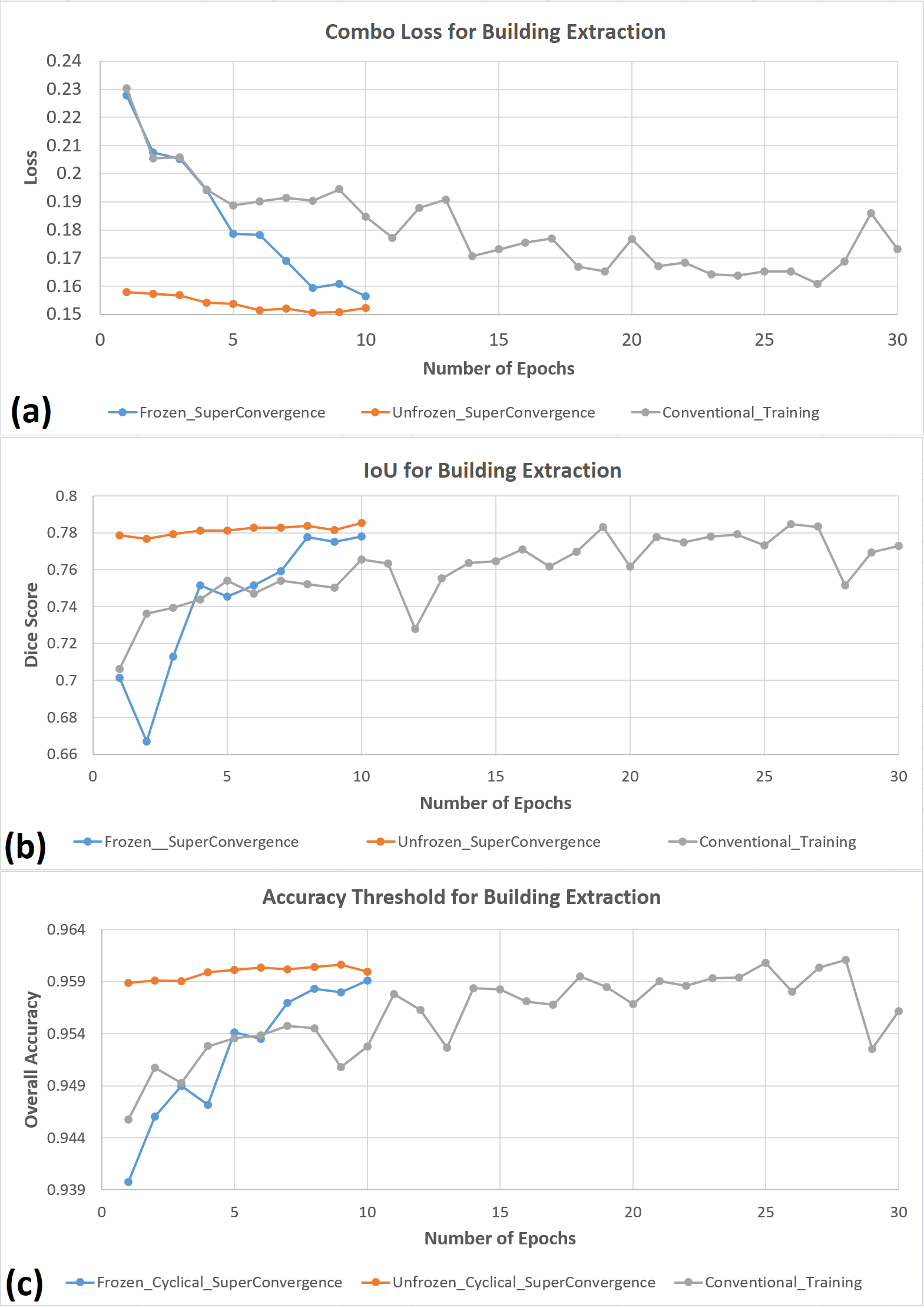}
  \caption{(a) The combo loss curves for three different training techniques for Res-U-Net trained on CB0 validation dataset, (b) The IoU curves for three different training techniques for Res-U-Net trained on CB0 validation dataset, (c) The accuracy curves for three different training techniques for Res-U-Net trained on CB0 dataset}
  \label{fig:clr_sc}
\end{figure}

Applying CLR and SC to fit the model while training with the frozen-unfrozen policy resulted in a very quick convergence of the model. The quicker convergence is favorable both in terms of combo loss (Equation 11), accuracy (Equation 12), and IoU (Equation 18). \autoref{fig:clr_sc}(a), (b) and (c) compare the combo loss, accuracy, and IoU of three training techniques -- frozen training with CLR and SC, unfrozen training with CLR and SC, and conventional training (without CLR or SC) -- on the same dataset (validation set of CB0 (R,G,B) band combination).

In all cases, it can be observed that conventional training policy takes more epochs (30) to converge. Frozen training with cyclical learning shows a steep convergence in the nearly one-third number of epochs (10) as compared to the conventional training policy. In the case of unfrozen training with cyclical learning, the starting accuracy and IoU are already very high, and the combo loss is very low, thanks to the frozen pretraining stage. The unfrozen policy eventually converges within 10 epochs with the apex of performance very close to the frozen policy. The proposed training methods were very efficient in terms of training time as well. \autoref{tbl-clr_sc} shows a performance comparison of the proposed training policy (Cyclical Learning, SuperConvergence, and Frozen-Unfrozen policies) and conventional training policy, on a machine with the following specifications: 8GB DDR4 RAM and 4 GB NVIDIA GTX 1050 Ti, with a clock speed of 3.5 GHz and four cores. It can be noted that the proposed training method performs better in terms of metrics and takes less than half the time to train, as compared to the conventional method.

\begin{table}
\caption{Comparison of Performance and Efficiency of Conventional and Proposed Training Policies that use CLR and SC}
\label{tbl-clr_sc}
\begin{tabular}{|l|l|l|}
\hline
                 & Conventional Training Policy & Proposed Training Policy (SC \& CLR) \\ \hline
Maximum Accuracy & 0.959                        & 0.962                                \\ \hline
Mean IoU         & 0.80                         & 0.80                                 \\ \hline
Mean F1-Score    & 0.88                         & 0.88                                 \\ \hline
Training Epochs  & 30                           & 10                                   \\ \hline
Training time    & $\sim$11.2 hours             & $\sim$4.5 hours                      \\ \hline
\end{tabular}
\end{table}

\subsection{Buildings under Shadows: A Typical Challenge}

\begin{figure}[h]
  \centering
  \includegraphics[width=01.0\linewidth]{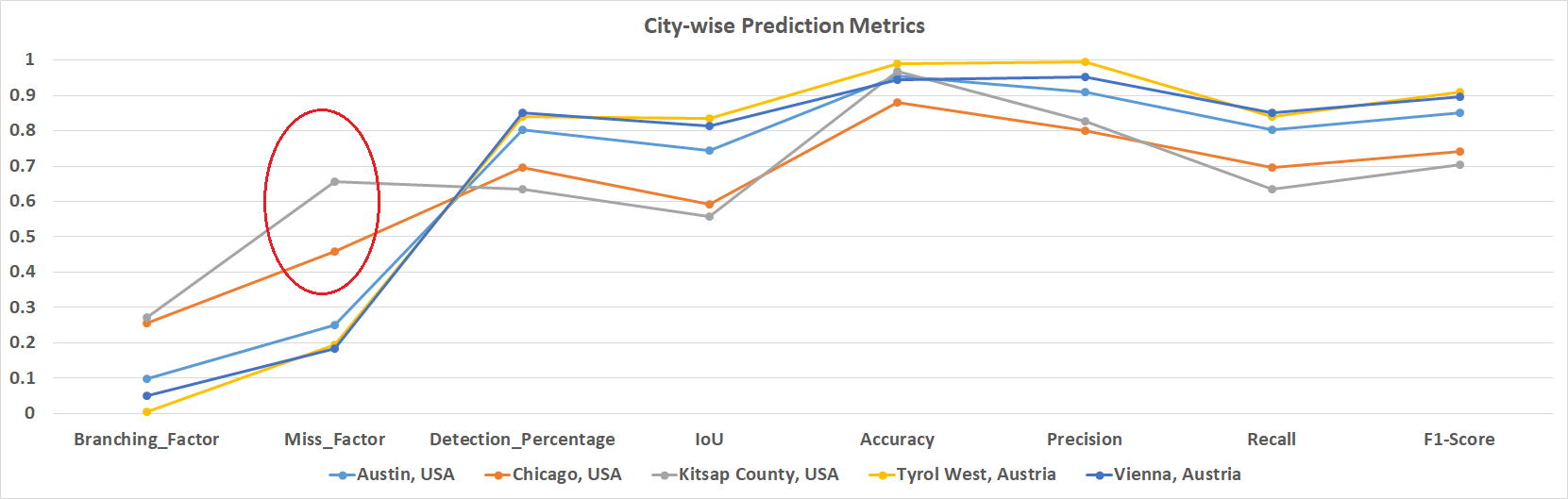}
  \caption{City-wise prediction metrics for the UAV test set, the high miss factor in Chicago and Kitsap is highlighted}
  \label{fig:shadow_issue_metrics}
\end{figure}

\begin{figure}
  \centering
  \includegraphics[width=01.0\linewidth]{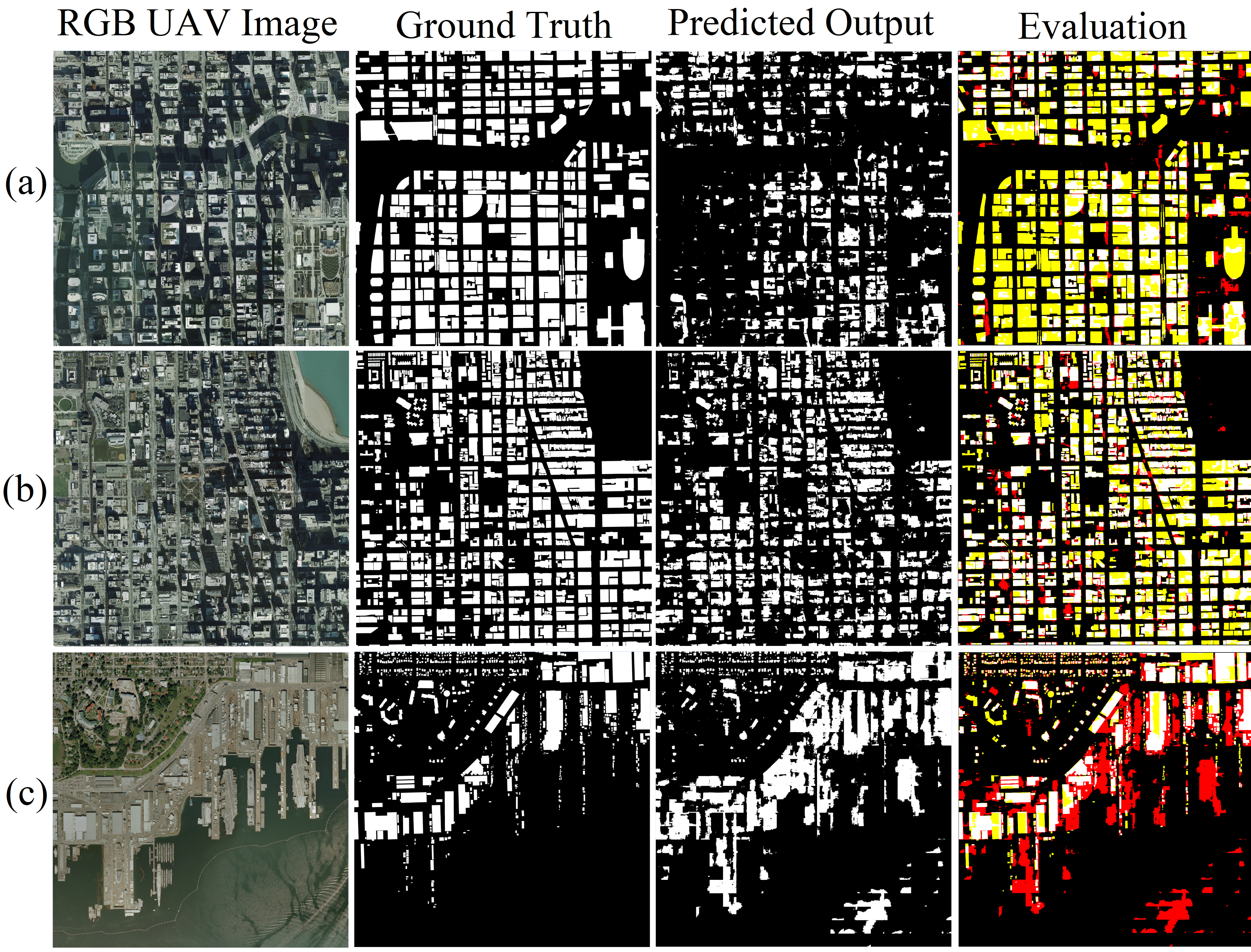}
  \caption{Select instances from the UAV test set of CB0 combination that led to high false positives and high miss factor. The first column is CB0 (RGB) input to the model, the second column is model prediction maps for building segmentation, the third column is ground truth, and the fourth column is evaluation images showing TP in white, TN in black, FP in red, and FN in yellow. (a), (b) From Chicago, USA (c) From Kitsap County}
  \label{fig:shadow_issue_segmentation}
\end{figure}

We performed a location-wise ablation analysis by plotting city-wise building segmentation metrics of the UAV test set, as shown in \autoref{fig:shadow_issue_metrics}. It can be observed that Chicago and Kitsap exhibited very high miss factors of over 40\% and 60\% respectively. Miss factor \autoref{eq-mf} is the ratio of the number of building pixels that our model missed to identify, to the number of building pixels that our model correctly identified and represents the omission error by the model. High miss factors (and hence high false negatives) for Chicago and Kitsap indicate that the model missed identifying over 40\% and 60\% of the buildings respectively for those two locations. Upon quantitative analysis of all the image instances from those two locations, we discovered that in most of those images, significant number of building pixels were under a shadow. Extracting buildings under shadows has been a prominent challenge for building segmentation \cite{Yang2023, Gao2018, Huang2012, Wang1993}. We selected three such shadowy instances from Chicago and Kitsap from the CB0 band combination (RGB) and used the proposed model to segment buildings as shown in \autoref{fig:shadow_issue_segmentation}. The evaluation maps for these instances showed a high number of false negatives, bumping up their miss factor. However, the model is successfully able to discriminate between buildings and shadows. The reasons for this could both be data-dependent and model-dependent. When an object is under a shadow, the loss of contrast also results in the loss of spectral variance as well as distinct spatial features such as edges. This can be observed in \autoref{fig:shadow_issue_segmentation}(a) and (b). Apart from shadows, \autoref{fig:shadow_issue_segmentation}(c), wherein over 30\% of water pixels are misidentified as building pixels, shows how a spectrally similar class can affect building segmentation performance. One of the learning representations for building segmentation is bright rooftops, based on high reflectance. It can be noted that these pixels represent brighter parts of the waterbody having high DN values due to saturation or sun-glint. In this case, the model was not able to enforce shape and edge restrictions, and misidentified water pixels as buildings.

\subsection{Comparison with Existing Benchmarks for Building Extraction}

Despite the challenges discussed in previous subsections and rare instance segmentation issues, the overall performance of the proposed dynamic Res-U-Net architecture with CB0 (R,G,B) band combination is highly favorable. Evaluation metrics, especially IoU, show that the while proposed model may face challenges for buildings under shadows, it successfully segments the buildings preserving edges as well as irregular shapes. Table 9 compares the best-case mean metrics of CB0 (R,G,B) in our study with other related studies that use at least one dataset from our combination of 5 datasets. The evaluation results from different studies shown in Table 9 are not directly comparable since each of them uses different deep-learning approaches and slightly different building footprint segmentation datasets. However, they give a general sense of how difficult the building footprint segmentation task is. We can see that our study provides a comparable model performance with state-of-the-art building segmentation methods.

\begin{table}
\caption{Comparison of proposed feature-guided Res-U-Net on CB0 (R,G,B) with other benchmark methods that use at least one of the 5 datasets that we combined, for building segmentation performance}
\label{tbl-model_benchmark_comparison}
\begin{tabular}{|l|l|l|l|l|l|}
\hline
Method                                          & Accuracy & Precision & Recall         & F1-Score & IoU   \\ \hline
Multiscale Res-U-Net (this paper)               & 0.965    & 0.883     & \textbf{0.861}  & \textbf{0.860}    & \textbf{0.80}  \\ \hline
Attention/Pyramid Networks \cite{Tian2022}      & 0.967    & -         & -              & -        & 0.768 \\ \hline
ResNet/DenseNet  \cite{Sariturk2023}            & 0.945    & 0.861     & 0.797          & 0.815    & 0.710 \\ \hline
Uni-scale Res-U-Net \cite{Dixit2021}            & 0.912    & 0.823     & 0.764          & 0.792    & 0.745 \\ \hline
HA-U-Net \cite{Xu2021}                          & 0.959    & -         & -              & 0.874    & 0.745 \\ \hline
Dilated CNNs \cite{khosh2020multiscale}         & 0.928    & -         & -              & 0.830    & 0.710 \\ \hline
\end{tabular}
\end{table}

\section{Conclusion}

This study aims at improving upon building segmentation from spectrally restricted, high spatial resolution remotely sensed imagery. A multiscale dataset is curated from 5 open-sources of RGB aerial and satellite data for training and validation. To expand the building segmentation performance within RGB channels, guiding features such as PC1, Sobel edge maps, VDVI, and MBI are extracted from the RGB channels. Using the RGB channels and the extracted features, three composite band combinations are defined namely Composite-0 (RGB), Composite-1(R:Sobel, G:VDVI, B:PC1), and Composite-2(R:Sobel, G:VDVI, B:MBI). A dynamic Res-U-Net architecture is then trained on each of these composites separately for building segmentation. A unique Combo-Loss is utilized to improve building segmentation specifically for irregular shapes as well as blurred boundaries. Training optimization techniques such as cyclical learning and SuperConvergence are employed, which reduce the training time to one-third of the conventional training methods. Evaluation of the proposed method across all three band combinations CB0, CB1, and CB2 reveals that in this study, although guiding features may not have improved the performance overall, they helped significantly in extracting buildings for instances having atypical backgrounds and/or landcover in the surrounding. The CB0 test set from UAV gives the best results, which are comparable with existing benchmarks for building segmentation. Moreover, to demonstrate the multiscale capability of the network, the model is also tested on an out-sample satellite image of the Chandigarh area, from the WorldView-3 sensor, and favorable results are obtained. One significant challenge that is identified in this study, both for UAV and satellite data is extracting the buildings lying under shadows. Almost all current methods are prone to high false negatives whenever a target building is under a shadow. This can be a potential future research question in the domain of building segmentation.

\section*{Data and Code Availability}
A python-based implementation of this project is available here: \url{https://github.com/Chintan2108/Object-Oriented-Deep-Learning-for-Building-Extraction}. To request the training data, please contact the first author.

\bibliographystyle{unsrt}  
\bibliography{mendeley}

\end{document}